\documentclass[letterpaper, 10 pt, conference]{ieeeconf}  

\IEEEoverridecommandlockouts                              

\usepackage{cite}
\usepackage{amsmath,amssymb,amsfonts}
\usepackage{algorithmic}
\usepackage{graphicx}
\usepackage{textcomp}
\usepackage{cite}
\usepackage{multirow}
\usepackage{array}
\usepackage{hyperref}
\usepackage[font=small]{caption}
\usepackage{subcaption}
\usepackage{balance}
\usepackage{subcaption}  
\usepackage{tabularx}
\usepackage[table]{xcolor}
\usepackage{multirow}
\usepackage[normalem]{ulem}
\useunder{\uline}{\ul}{}

\usepackage{graphicx}

\title{\LARGE \bf
UAV-CodeAgents: Scalable UAV Mission Planning via Multi-Agent ReAct and Vision-Language Reasoning}

\author{Oleg Sautenkov, Yasheerah Yaqoot, Muhammad Ahsan Mustafa, Faryal Batool, \\ Jeffrin Sam, Artem Lykov, Chih-Yung Wen\textsuperscript{\dag}, and Dzmitry Tsetserukou
\thanks{The authors are with the Intelligent Space Robotics Laboratory, Center for Digital Engineering, Skolkovo Institute of Science and Technology ({\tt \{oleg.sautenkov, yasheerah.yaqoot, ahsan.mustafa, faryal.batool, jeffrin.sam, artem.lykov, d.tsetserukou\}@skoltech.ru}) and \textsuperscript{\dag} with the AiRo Lab, Department of Aeronautical and Aviation Engineering, Hong Kong Polytechnic University ({\tt chihyung.wen@polyu.edu.hk}).}
}

\begin{document}

\maketitle
\thispagestyle{empty}
\pagestyle{empty}


\begin{abstract}

We present \textbf{UAV-CodeAgents}, a scalable multi-agent framework for autonomous UAV mission generation, built on large language and vision-language models (LLMs/VLMs). The system leverages the ReAct (Reason + Act) paradigm to interpret satellite imagery, ground high-level natural language instructions, and collaboratively generate UAV trajectories with minimal human supervision. A core component is a vision-grounded, pixel-pointing mechanism that enables precise localization of semantic targets on aerial maps. To support real-time adaptability, we introduce a reactive thinking loop, allowing agents to iteratively reflect on observations, revise mission goals, and coordinate dynamically in evolving environments.

UAV-CodeAgents is evaluated on large-scale mission scenarios involving industrial and environmental fire detection. Our results show that a lower decoding temperature (0.5) yields higher planning reliability and reduced execution time, with an average mission creation time of 96.96 seconds and a success rate of 93\%. We further fine-tune Qwen2.5VL-7B on 9,000 annotated satellite images, achieving strong spatial grounding across diverse visual categories. To foster reproducibility and future research, we will release the full codebase and a novel benchmark dataset for vision-language-based UAV planning. 

\end{abstract}

\section{Introduction}

Autonomous UAV systems are increasingly tasked with complex missions that demand both high-level understanding and spatial precision—ranging from large-scale environmental assessment to time-critical operations in dynamic terrains \cite{ai_meets_uav}, \cite{uav_survey}. Traditional UAV planning pipelines rely heavily on predefined maps, hand-engineered heuristics, or manual waypoint configurations \cite{sautenkov2024flightararflightassistance}, limiting their adaptability and scalability. Recent progress in multimodal AI—specifically Large Language Models (LLMs) and Vision-Language Models (VLMs)—offers a promising alternative by enabling agents to reason about unstructured language and image inputs. However, current frameworks typically operate in closed environments or single-agent settings, and fail to harness the full potential of collaborative reasoning or grounded spatial planning.

This paper introduces \textbf{UAV-CodeAgents}, a scalable multi-agent framework for autonomous mission generation in geospatial environments. Built on the ReAct (Reason + Act) paradigm, our agents are equipped to interpret satellite imagery, ground language instructions at the pixel level, and iteratively revise plans based on environmental cues and prior decisions. The system enables agents to localize semantic targets on aerial maps, extract spatial goals, and generate context-aware flight routes with minimal human supervision. This grounding capability forms the foundation for high-level autonomy, where agents are not just following paths but understanding what and why to explore.

\begin{figure}[t]
    \flushright
    \includegraphics[width=0.5\textwidth]{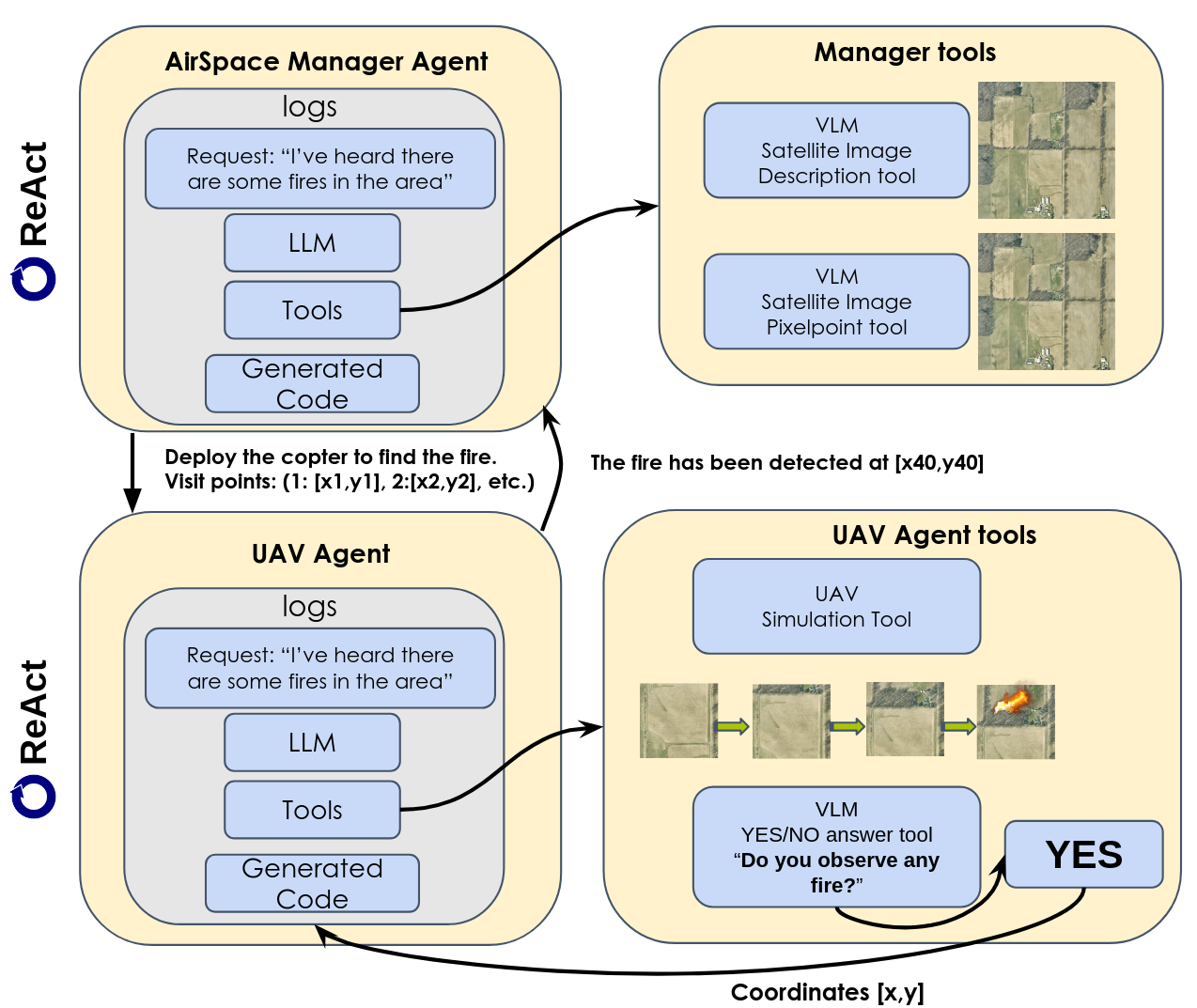}
    \caption{The pipeline of the UAV-CodeAgents system.}
    \label{main_pic}
\end{figure}

At the core of UAV-CodeAgents is \textit{reactive thinking} loop (ReAct), which enables agents to reflect on ambiguous or incomplete information, re-query the visual scene, and adjust their actions dynamically. This allows the system to generalize to diverse scenarios, adapt to mission constraints, and operate effectively in partially known environments, bridging the gap between large-scale reasoning and low-level UAV control.

\vspace{1mm}
\noindent The main contributions of this work are:
\begin{itemize}
    \item We propose \textbf{UAV-CodeAgents}, a multi-agent framework that integrates LLMs and VLMs under the ReAct paradigm for collaborative, vision-language-driven UAV mission generation.
    \item We release a \textbf{benchmark} for evaluating semantic grounding and collaborative planning from aerial imagery and textual prompts, enabling reproducible research in scalable UAV autonomy.
    \item We introduce \textbf{new dataset}, aimed at precision pixelpointing on the satellite images.
    \item We design a UAV-oriented \textbf{reactive thinking loop} that enables iterative visual reasoning, uncertainty resolution, and plan refinement, allowing agents to adapt in dynamic mission settings.
    \item We finetuned the \textbf{Qwen2.5VL-7B} on a novel dataset, that aligns semantic language entities with coordinates on satellite imagery for precise waypoint extraction.
    \item We design a UAV-oriented \textbf{reactive thinking loop} that enables iterative visual reasoning, uncertainty resolution, and plan refinement, allowing agents to adapt to the mission settings.
\end{itemize}


\section{Related Work}
\label{sec:literature}

In recent years, Vision-Language-Action (VLA) and Vision-Language (VLM) systems have emerged as a transformative approach in robotic autonomy, particularly for Unmanned Aerial Vehicles (UAVs). By combining perception, cognition, and control through natural language and visual inputs, these systems offer promising capabilities for high-level reasoning, mission planning, and real-time decision-making. This section reviews foundational models, recent advances in UAV-specific systems, and innovations in multi-agent reasoning and retrieval-augmented frameworks.

\subsection{Foundational Vision-Language Models and Navigation}

The development of transformer-based vision-language models laid the groundwork for VLA research. Dosovitskiy et al.'s Vision Transformer (ViT) \cite{visiontransformerdosovitskiy2021imageworth16x16words} and CLIP \cite{CLIPradford2021learningtransferablevisualmodels} demonstrated scalable multimodal representation learning, while models like GPT-4o \cite{openai2024gpt4technicalreport} enabled unified vision, language, and audio reasoning. These architectures inspired early UAV-focused systems such as AerialVLN \cite{liu2023aerialvlnvisionandlanguagenavigationuavs}, which enabled drones to interpret natural language instructions, and dialog-based extensions \cite{fan2023aerialvisionanddialognavigation} that improved UAV-human interaction.

To simulate aerial autonomy under realistic conditions, Wang et al. proposed the OpenUAV benchmark \cite{wang2024realisticuavvisionlanguagenavigation}, while Gao et al. developed Semantic-Topo-Metric Representations (STMR) \cite{gao2024aerialvisionandlanguagenavigationsemantictopometric} to enhance UAVs’ spatial understanding. EmbodiedCity \cite{EmbodiedCityzhang2024} extended these efforts to city-scale VLN simulations, leveraging real-world pedestrian and vehicle data for urban navigation. Karaf et al. \cite{karaf2025morphonaviaerialgroundrobotnavigation} used foundation models to guide legged aerial robots in complex environment.

\subsection{Cognitive Reasoning and Planning in UAVs}

While foundational VLA systems focus on perception and basic control, more recent efforts incorporate cognitive reasoning to handle symbolic understanding and logical decision-making. CognitiveDrone \cite{lykov2025cognitivedronevlamodelevaluation} introduced a two-tier VLA architecture with a dedicated reasoning module, achieving a 77.2\% success rate on cognitive UAV tasks such as human recognition and symbol reasoning. These findings underscore the importance of decoupling high-level cognition from low-level control for real-time autonomy.

Complementary efforts in mission planning further extend these ideas. UAV-VLA \cite{sautenkov2025uavvlavisionlanguageactionlargescale} and UAV-VLPA* \cite{sautenkov2025uavvlpavisionlanguagepathactionoptimalroute} combined vision-language understanding with trajectory optimization using A* and TSP solvers. UAV-VLRR \cite{yaqoot2025uavvlrrvisionlanguageinformednmpc} applied these capabilities to onboard Nonlinear Model Predictive Control (NMPC) for real-time, cluttered environment navigation. Meanwhile, RaceVLA \cite{serpiva2025racevlavlabasedracingdrone} showcased the feasibility of end-to-end velocity and yaw prediction directly from natural language and visual inputs.

\subsection{Scalability, Multi-Agent Reasoning, and Retrieval-Augmented Generation}

Scalability and open-world deployment have become critical research directions. Open-source models like OpenVLA \cite{kim2024openvlaopensourcevisionlanguageactionmodel} and MiniVLA \cite{belkhale2024minivla} support extensibility and community collaboration. Multi-agent coordination has also gained traction with frameworks like SwarmGPT \cite{jiao2023swarmgptcombininglargelanguage} and FlockGPT \cite{lykov2024flockgptguidinguavflocking}, which demonstrate LLM-driven behavior orchestration and SDF-based spatial strategies for swarm UAVs.

In parallel, Retrieval-Augmented Generation (RAG) has emerged as a powerful mechanism for embedding domain knowledge into LLM-based agents. WildfireGPT \cite{xie2025wildfiregpttailoredlargelanguage} exemplifies this paradigm by integrating a modular reasoning agent for wildfire risk analysis. It leverages user profiling, memory, and strategic planning modules to refine decision-making with climate data and academic literature. UAV-CodeAgents builds upon these ideas by introducing an iterative ReAct-style reasoning framework for UAVs, enabling pixel-level geospatial grounding and dynamic mission adaptation across multiple agents.

\section{System Overview}
\label{sec:systemoverview}

UAV-CodeAgents is a modular and scalable framework designed for autonomous mission generation in aerial robotics through vision-language reasoning and collaborative multi-agent planning. The architecture follows a decentralized paradigm, composed of heterogeneous agents specialized in reasoning, perception, and physical interaction with the environment. By leveraging the ReAct (Reason + Act) model and a novel pixel-level vision-language interface, UAV-CodeAgents enables efficient interpretation of vague or high-level instructions into executable UAV trajectories across complex terrains.

\begin{table*}[h]
    \centering
    \footnotesize
    \renewcommand{\arraystretch}{1.3}
    \setlength{\tabcolsep}{8pt}
    \caption{\textsc{Tools and Functionalities of CodeAgent and UAV Agent}}
    \label{tab:agent_tools}
    \begin{tabularx}{\textwidth}{|l|>{\ttfamily}l|X|}
    \hline
    \rowcolor[HTML]{e6e6e6}
    \textbf{Agent} & \textbf{Tool Name} & \textbf{Description and Arguments} \\
    \hline
    \multirow{5}{*}{\textbf{Airspace Manager Agent}} 
      & read\_image & Returns a PIL image object for airspace analysis. \\
      & & \textit{Arguments:} \texttt{i} (integer) - image index \\
      \cline{2-3}
      & describe\_satellite\_image & Analyzes image using Qwen2.5-VL via Fireworks API. \\
      & & \textit{Arguments:} \texttt{image} (PIL.Image) - image to process \\
      \cline{2-3}
      & pixelpoint\_objects & Extracts objects with Qwen2.5-VL, auto-repairs JSON. \\
      & & \textit{Arguments:} \texttt{image} (PIL.Image), \texttt{objects} (string) \\
      \cline{2-3}
      & visualize\_keypoints & Renders labeled objects on image with bounding boxes. \\
      & & \textit{Arguments:} \texttt{image} (PIL.Image), \texttt{keypoints} (list) \\
      \cline{2-3}
      & final\_answer & Delivers processed results to user. \\
      & & \textit{Arguments:} \texttt{answer} (any type) \\
    \hline
    \multirow{4}{*}{\textbf{UAV Agent}}
      & read\_image\_for\_simulation & Provides image for UAV flight simulation. \\
      & & \textit{Arguments:} \texttt{i} (integer) - simulation index \\
      \cline{2-3}
      & uav\_simulation & Simulates flight path and captures frames. \\
      & & \textit{Arguments:} \texttt{image} (PIL.Image), \texttt{labeled\_points} (list) \\
      \cline{2-3}
      & detect\_and\_display & Identifies fire locations in UAV footage. \\
      & & \textit{Arguments:} \texttt{frames\_dict} (dictionary) \\
      \cline{2-3}
      & final\_answer & Returns detection results. \\
      & & \textit{Arguments:} \texttt{answer} (any type) \\
    \hline
    \end{tabularx}
\end{table*}

\subsection{Architecture Overview}

The system is composed of the following core components, each fulfilling a critical role in the perception-to-action pipeline:

\subsubsection{ Airspace Management Agent (AMA)}
This is the central reasoning agent, responsible for interpreting natural language commands, analyzing satellite images, and generating spatially grounded mission plans. It interfaces with both language and vision models to produce semantically aligned geospatial outputs.

\begin{itemize}
\item \textbf{Hierarchical Task Decomposition:} Utilizes LLMs (e.g., GPT-4, Qwen2.5VL) to parse and deconstruct user input into structured tasks (e.g., search, localize, verify).
\item \textbf{Visual Scene Understanding:} Employs vision-language models to generate scene-level summaries and detect high-risk or mission-critical zones (e.g., fire, damaged structures).
\item \textbf{Pixel-Level Grounding:} Uses fine-tuned VLMs to identify coordinates on satellite imagery corresponding to natural language phrases, using attention-weighted pixel-pointing.
\item \textbf{Path Optimization:} Makes the path of UAV in order to visit the objects with the highest probabilty of fire first.
\end{itemize}

\subsubsection{ UAV Agent}
This module corresponds to the embodied UAVs (either simulated or real) that execute assigned missions. Each UAV agent is endowed with lightweight reasoning capability to allow real-time re-evaluation of mission steps and environment changes.

\begin{itemize}
\item \textbf{Autonomous Flight Performance:} Executes waypoint-following based on the planned route from the AMA.
\item \textbf{Image Acquisition and Inference:} Captures high-resolution RGB images during flight; performs VLM inference to update task status (e.g., fire confirmed, object located).
\item \textbf{Feedback and Reporting:} Sends observations, semantic summaries, and confidence scores back to the AMA for iterative refinement of the mission plan.
\end{itemize}

\subsection{Communication and Synchronization}

Our system is built on top of the smolagents framework, which provides a lightweight and modular foundation for multi-agent coordination. Agents communicate via a simple message-passing interface. Each agent periodically reports its state (position, image, semantic annotations). This architecture supports fault-tolerance and asynchronous operation.

\subsection{Reactive Reasoning Loop (ReAct)}
The ReAct paradigm is central to the system's robustness and adaptability. It enables agents to iteratively reflect on ambiguous or incomplete inputs, re-evaluate current assumptions, and revise their actions accordingly. This loop is instantiated as follows:

\begin{enumerate}
\item \textbf{Observe:} Capture the current scene via UAV image or satellite snapshot.
\item \textbf{Describe:} Use a VLM to generate a high-token-length scene caption.
\item \textbf{Reason:} Use the LLM to interpret the scene description in the context of the mission query.
\item \textbf{Decide:} Identify or update waypoints, assign new UAV roles, or revise task priorities.
\item \textbf{Act:} Deploy updated instructions to the relevant UAV agents.
\end{enumerate}

This enables dynamic behavior where, for instance, if a fire is suspected but not visually confirmed, a UAV may be redirected for closer inspection or re-imaging from a different angle.

\subsection{Pixel-Pointing Grounding Mechanism}

While not a novel method, pixel-pointing is a key component of UAV-CodeAgents, enabling precise localization of language-referenced entities (e.g., “the warehouse near the forest”) directly on satellite maps. To improve performance in this domain, we fine-tuned Qwen-VL-2.5-7B using Supervised Fine-Tuning (SFT) on a custom dataset of 9,000 annotated satellite images. This fine-tuning specifically targeted pixel-level grounding in aerial and satellite imagery.

\subsection{Scalability and Extensibility}

UAV-CodeAgents is designed to operate with flexible agent counts and varying computational budgets. It supports:

\begin{itemize}
\item \textbf{Plug-and-Play Agents:} New UAVs can be dynamically added or removed during missions.
\item \textbf{Heterogeneous Models:} Different agents may run different models (e.g., lightweight VLMs on drones, powerful LLMs in the AMA).
\item \textbf{Sim2Real Transfer:} A simulated testing layer enables safe benchmarking before real-world deployment.
\end{itemize}




\section{Case Study: Fire Detection}

\subsection{Benchmark}
The system's detection capabilities were quantitatively evaluated using an augmented version of the UAV-VLA benchmark\cite{sautenkov2025uavvlavisionlanguageactionlargescale}. Our evaluation framework incorporated 30 RGB images  spanning four critical fire scenarios:
Urban fires, Industrial fires, Wildland fires, Composite events, Vehicles.

\begin{figure}[htbp]
    \centering
    \begin{subfigure}{0.2\textwidth}
        \centering
        \includegraphics[width=1.1\linewidth]{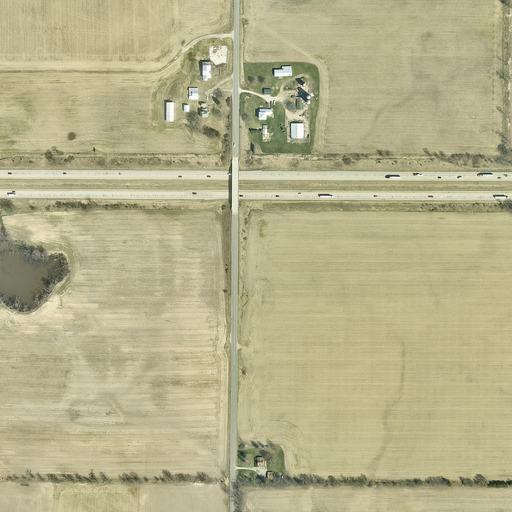}
        \caption{Normal conditions }
        \label{fig:bench_no_fire}
    \end{subfigure}
    \hspace{0.05\textwidth}  
    \begin{subfigure}{0.2\textwidth}
        \centering
        \includegraphics[width=1.1\linewidth]{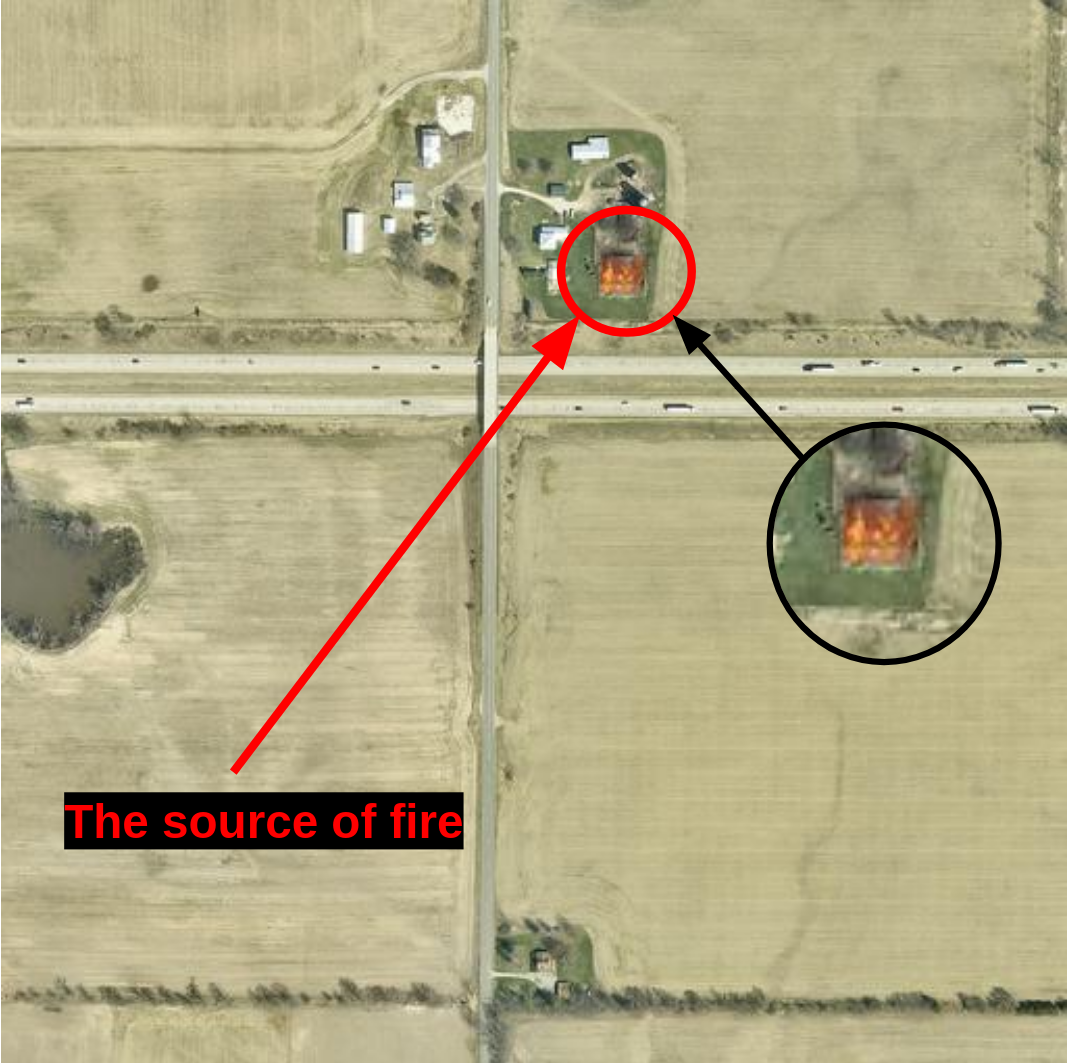}
        \caption{Fire incident scenario}
        \label{fig:bench_fire}
    \end{subfigure}
    \caption{The samples in the benchmark.}
    \label{fig:fire_comparison}
\end{figure}

The system was tested against ambiguous natural language query to evaluate its reasoning capabilities. The primary test phrase:
\textbf{"I've heard there are fires in our area."}
was selected to assess the system's ability to infer potential fire locations from vague descriptions and resolve spatial references through contextual understanding.

\subsection{Simulation}

\begin{figure}[h]
\includegraphics[width=0.4\textwidth]{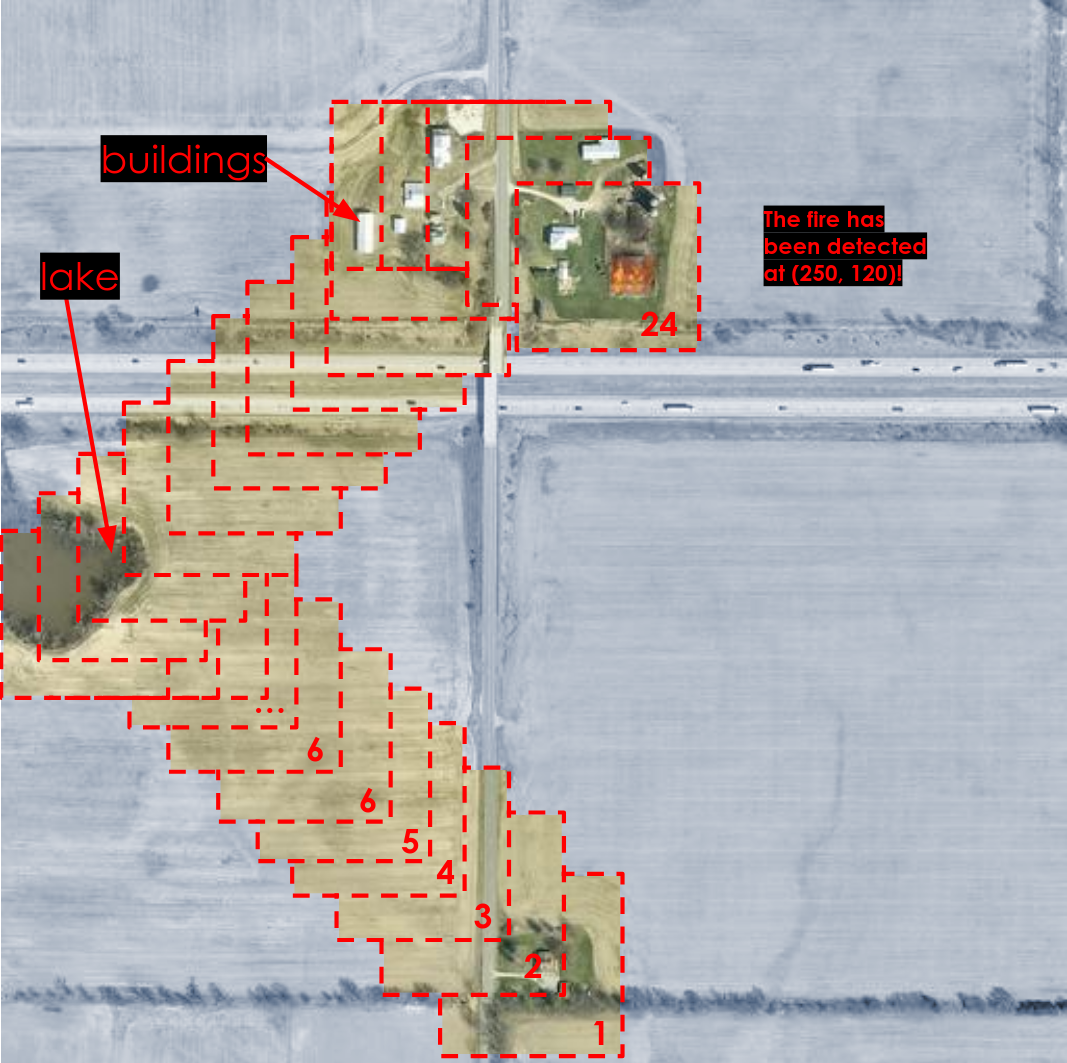}
    \caption{The scheme of simulation.}
    \label{Simulation}
    \centering
\end{figure}

To visualize and test UAV navigation and perception capabilities, we implemented a lightweight 2D simulation environment. The simulator emulates UAV flight over satellite or aerial imagery by interpolating movement across a user-defined sequence of labeled waypoints. Each waypoint consists of a 2D pixel coordinate and an associated semantic label (e.g., ``lake'', ``building'', ``road'').

The simulator performs the following steps:

\begin{itemize}

\item \textbf{Trajectory Interpolation:} For each pair of consecutive waypoints, the system computes a series of intermediate UAV positions using linear interpolation. This models smooth, continuous UAV flight.

\item \textbf{View Cropping and Annotation:} At each interpolated position, a crop of the scene centered on the UAV's location is extracted. Each crop is then resized and annotated with the corresponding label. This mimics the UAV's field of view and real-time annotation of observed regions.

\item \textbf{Frame Collection and GIF Generation:} All generated image crops are compiled into an animated GIF, providing a visual trace of the UAV’s flight path across the mission. This output is displayed inline for rapid iteration and evaluation.

\item \textbf{Output for Further Processing:} In addition to the visualization, the simulator returns a dictionary mapping frame identifiers to tuples of (cropped image, center coordinates). This structure supports downstream tasks such as behavior analysis, training data generation, or agent feedback.

\end{itemize}

This simulation component allows us to rapidly prototype navigation behaviors, evaluate perception outputs, and generate visual demonstrations of UAV agent decisions without the overhead of a full physics-based environment.
Complete quantitative results are presented in Section~\ref{sec:result}.
	
\section{Experimental Analysis}
\label{sec:result}

\subsection{Experimental Setup}
We conducted our experiments using models from the Qwen family, selected for their strong performance in both language understanding and multimodal reasoning tasks.

For the core decision-making components of our system—namely, the Airspace Manager Agent (AMA) and the UAV Agent—we employed the Qwen2.5-72B model. This model was responsible for high-level task planning, inter-agent coordination, and dynamic mission updates during runtime.

To handle visual perception tasks, such as identifying and describing fire zones from satellite imagery, we integrated Qwen2.5VL-32B, accessed via the Fireworks API.

We evaluated model behavior at two different sampling temperatures—0.5 and 0.7—to analyze the trade-off between determinism and creativity in decision-making. Temperature 0.5 was generally preferred for more stable and consistent reasoning in flight coordination, while temperature 0.7 was explored to allow more flexible and diverse outputs in image interpretation tasks.

\subsection{Performance Metrics}
Detection performance was measured using the following metric:
\begin{equation}
\text{Time-to-Detection (TTD)} = \frac{1}{N}\sum_{i=1}^{N}(t_{\text{detect}}^{(i)} - t_{\text{query}}^{(i)}),
\end{equation}
where $N = 30$ test cases, excluding those with false positives or false negatives.

\subsection{Experimental Results}

In general, lower temperature settings (e.g., 0.5) were found to be preferable for the base model, as they led to better consistency, faster execution, and higher overall performance. The model at temperature 0.7 not only produced fewer successful samples \textbf{(26 vs. 28)} but also resulted in longer average task durations \textbf{(105.29 s vs. 96.96 s)}, as shown in Table~\ref{tab:average_time}.


\begin{table}[h!]
\caption{\textsc{Effect of Sampling Temperature on Mission Execution Time and Success Rate}}
\centering
\begin{tabular}{|c|c|c|}
\hline
\textbf{\begin{tabular}[c]{@{}c@{}}Sampling\\ Temperature (T)\end{tabular}} & \textbf{\begin{tabular}[c]{@{}c@{}}Avg. Elapsed \\ Time (s)\end{tabular}} & \textbf{Successful samples} \\ \hline
0.5                                                                         & 96.96                                                                     & 28                          \\ \hline
0.7                                                                         & 105.29                                                                    & 26                          \\ \hline
\end{tabular}
\label{tab:average_time}
\end{table}

Furthermore, the 0.7 setting struggled particularly with the following samples: 6, 23, 24, and 29, where it failed to generate coherent plans or misinterpreted image-grounded inputs. These inconsistencies suggest that the added randomness at higher temperatures can negatively impact the reliability of UAV coordination, especially in mission-critical scenarios.

\begin{figure}[h]
    \centering
    \includegraphics[width=0.48\textwidth]{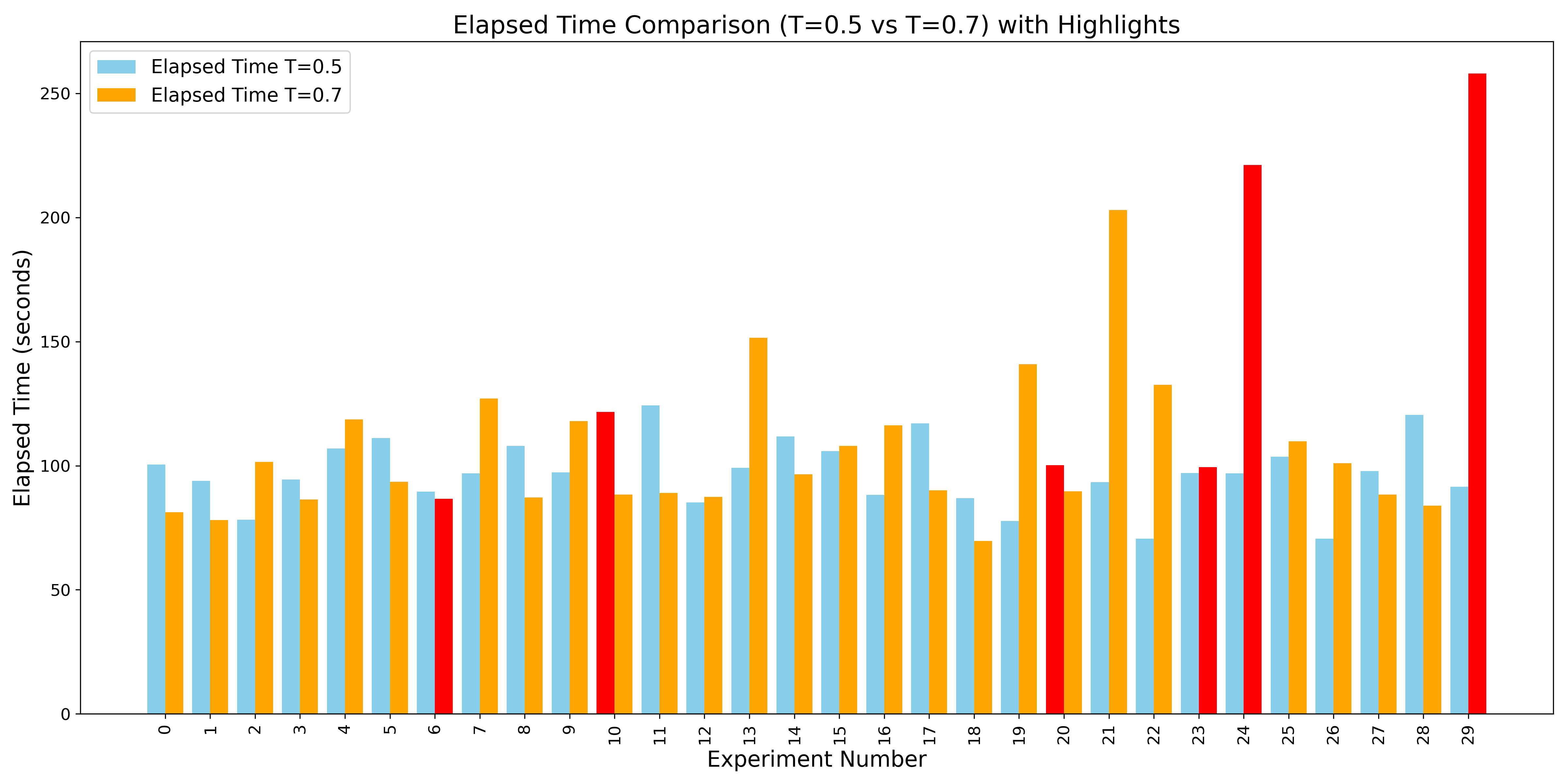}
    \caption{Task completion times across temperature settings.}
    \label{evaluation_time}
\end{figure}

\subsection{Experimental Results of SFT for Qwen2.5VL-7B}

\begin{figure}[h]
    \centering
    \includegraphics[width=0.48\textwidth]{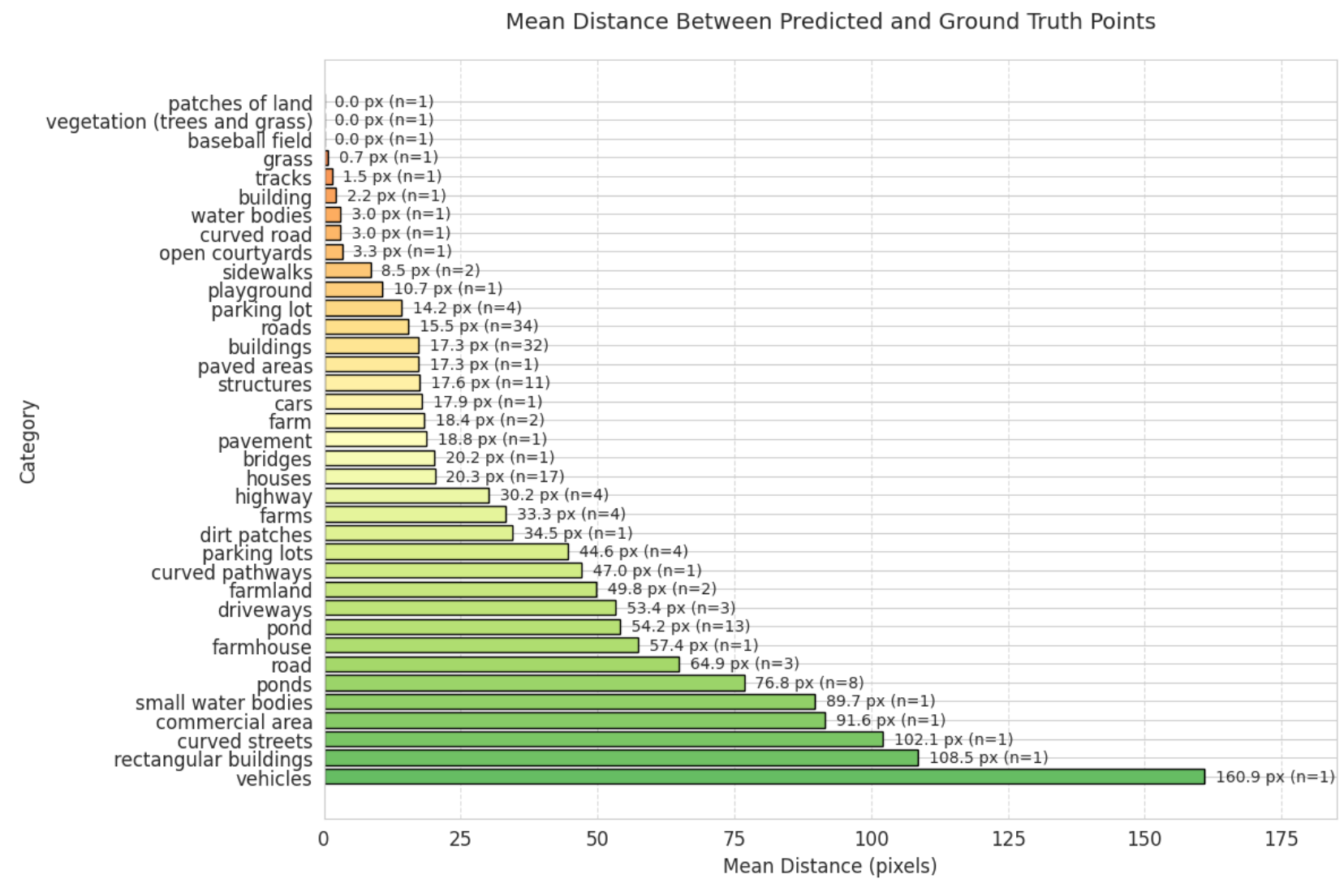}
    \caption{Mean distance in the pixel-pointing task.}
    \label{finetuning_data}
\end{figure}

\begin{table}[h!]
\centering
\caption{\textsc{Summary of Mean Distance and Mean Coverage by Category}}
\begin{tabular}{|l|r|r|}
\hline
\textbf{Category} & \textbf{Mean Distance} & \textbf{Mean Coverage} \\
\hline
ponds         & 76.77 & 1.22 \\
road          & 64.93 & 1.63 \\
pond          & 54.18 & 3.08 \\
driveways     & 53.35 & 3.57 \\
farmland      & 49.80 & 2.55 \\
parking lots  & 44.56 & 2.41 \\
farms         & 33.28 & 0.74 \\
highway       & 30.17 & 1.72 \\
houses        & 20.33 & 1.32 \\
farm          & 18.36 & 1.69 \\
structures    & 17.58 & 0.89 \\
buildings     & 17.31 & 1.54 \\
roads         & 15.45 & 1.21 \\
parking lot   & 14.18 & 0.81 \\
sidewalks     & 8.52  & 0.86 \\
\hline
\end{tabular}
\end{table}

The model demonstrates strong semantic grounding across a wide range of visual categories such as ponds, roads, parking lots, and buildings. These categories exhibit moderate to high mean distances—indicating distributed object presence—and non-trivial coverage values, suggesting effective spatial understanding.

Categories like ``pond'' (mean coverage = 3.08), ``driveways'' (3.57), and ``farmland'' (2.55) indicate that the model frequently identifies multiple or extended instances of relevant content per sample, reflecting a broad contextual awareness. Conversely, categories such as ``sidewalks'', ``structures'', and ``parking lot'' show lower coverage values (0.8–0.9) with moderate distances, pointing to fine-grained and localized object detection.

Overall, the fine-tuned Qwen2.5VL-7B model performs well across both sparse and dense scene elements, achieving reliable pixel-level localization. Its context-aware grounding makes it suitable for downstream applications in aerial image analysis, urban planning, and autonomous UAV navigation.



\section{Conclusion and Future Work}
\label{sec:conclusion}

This work presents \textbf{UAV-CodeAgents}, a scalable, vision-language-guided multi-agent system for autonomous UAV mission generation. Built upon the smolagents framework and powered by Qwen2.5-based LLMs and VLMs, the system enables decentralized reasoning, semantic grounding at the pixel level, and adaptive planning through a reactive thinking loop. UAV-CodeAgents bridges the gap between high-level instruction following and low-level geospatial execution in real-time, multi-agent settings.

\vspace{1mm}
Our study demonstrates that system can effectively coordinate UAV operations using large language and vision-language models. By integrating a ReAct-style reasoning loop, agents are able to revise plans iteratively, allowing for greater adaptability in dynamic or uncertain environments.  Experimental results show that a lower sampling temperature of 0.5 leads to more consistent and efficient performance, with the system successfully processing 28 out of 30 images—achieving a 93\% success rate with an average completion time of 96.96 seconds. This highlights the reliability of the system under low-temperature settings compared to higher-temperature ones, which demonstrated slower and less stable behavior. The architecture’s modularity allows for scalable deployment, flexible UAV integration, and potential for sim2real transfer. Looking ahead, we aim to extend CodeAgents with swarm of UAVs, real-time telemetry integration, and support for sensors, moving toward robust swarm coordination in real-world scenarios like disaster response and environmental monitoring.


\balance

\bibliographystyle{IEEEtran}
\bibliography{ref}

\end{document}